%% file: root.tex
\documentclass[letterpaper, 10 pt, conference]{ieeeconf}  

\makeatletter
\let\MATCHclassauthor\@author
\let\MATCHclasscaption\caption
\let\MATCHclassatcaption\@caption
\let\MATCHclassmakecaption\@makecaption
\newskip\MATCHabovecaptionskip
\newskip\MATCHbelowcaptionskip
\MATCHabovecaptionskip=\abovecaptionskip
\MATCHbelowcaptionskip=\belowcaptionskip
\makeatother

\IEEEoverridecommandlockouts                              

\usepackage{graphics}
\usepackage{amsmath}
\usepackage{amssymb}
\usepackage[colorlinks, linkcolor=blue, citecolor=blue]{hyperref}
\usepackage{cite}
\usepackage{algorithm}
\usepackage{url}
\usepackage{algpseudocode}
\usepackage{epsfig}
\usepackage{graphicx}
\usepackage{caption}
\usepackage{subfigure}
\usepackage{mathrsfs}
\usepackage{array}
\usepackage{multirow}
\usepackage{multicol}
\usepackage{booktabs}
\usepackage{makecell}
\usepackage{xcolor}
\usepackage{color}
\usepackage{amsfonts}
\usepackage{gensymb}
\usepackage{threeparttable}
\usepackage{textcomp}

\usepackage{bibunits}
\usepackage{docmute}
\makeatletter
\AtBeginDocument{%
  \let\MATCHstd@bibitem\std@bibitem
  \def\std@bibitem#1{%
    \MATCHstd@bibitem{#1}%
    \Hy@raisedlink{\hyper@anchorstart{cite.#1\@extra@b@citeb}\hyper@anchorend}%
    \ignorespaces}%
  \renewcommand\bibcite[2]{%
    \global\@namedef{b@#1\@extra@b@citeb}{%
      \hyper@@link[cite]{}{cite.#1\@extra@b@citeb}{#2}}}}
\makeatother

\IEEEaftertitletext{\vspace{-1em}}

\title{\LARGE \bf
MATCH: Execution-Aware Hierarchical Reinforcement Learning for Vision-and-Language Navigation in Continuous Environment
}

\author{
Shuhao Ye$^{1}$, Yufei Wei$^{1}$, Demao Wang$^{1}$, Shichao Zhai$^{1}$, Rong Xiong$^{1}$, Yue Wang$^{1}$$^{\dagger}$
\thanks{$^{1}$Zhejiang University, Hangzhou, China}%
\thanks{$\dag$Corresponding author: Yue Wang (email: wangyue@iipc.zju.edu.cn)
}
}

\begin{document}
\makeatletter
\let\MATCHmaketitle\maketitle
\let\MATCHatmaketitle\@maketitle
\def\@extra@b@citeb{.main}
\makeatother
\begin{bibunit}[ieeetr]

\maketitle
\thispagestyle{empty}
\pagestyle{empty}

\begin{abstract}
Vision-and-Language Navigation in Continuous Environments (VLN-CE) requires instruction-consistent planning and reliable execution under collision constraints. Yet many hierarchical approaches optimize the planner while keeping execution fixed. We introduce MATCH, a framework that jointly learns a language-conditioned planner and a goal-conditioned primitive-action executor. An execution-aware topological graph represents start-dependent execution options and integrates executor value estimates with realized outcome memory, enabling the planner to select execution requests and refine their targets through bounded continuous offsets. A transition-aware dense macro reward balances ordered progress along the reference route against execution residual, encouraging executable requests without substituting local success for instruction following. A staged curriculum culminates in synchronized on-policy updates, with the executor learning on planner-generated requests and the planner adapting its requests to the executor’s current behavior. With 0.5B parameters, MATCH achieves state-of-the-art performance on the R2R-CE and RxR-CE benchmarks. Preliminary real-robot deployments further demonstrate its practical potential.
\end{abstract}

\section{Introduction}

Vision-and-Language Navigation in Continuous Environments (VLN-CE)~\cite{krantz2020beyond} requires an agent to follow natural-language instructions through hundreds of collision-constrained primitive actions.
Many approaches address this long-horizon problem by predicting intermediate navigation intents, such as target locations~\cite{an2024etpnav, wei2026dualvln, ye2026revisiting}, or short waypoint-based trajectories~\cite{zhang2026navfom, xue2026omninav}, which must then be translated into movement.
We refer to intent selection as macro-level planning and intent realization as micro-level execution.
In the hierarchical Markov decision process (HMDP) view~\cite{parr1998hmdp}, an intent is a macro action whose state transition depends on both environmental dynamics and the executor's policy.
Selecting an instruction-consistent intent therefore doesn't ensure that it can be realized under collision constraints.

\begin{figure}[t]
    \centering
    \includegraphics[width=\linewidth]{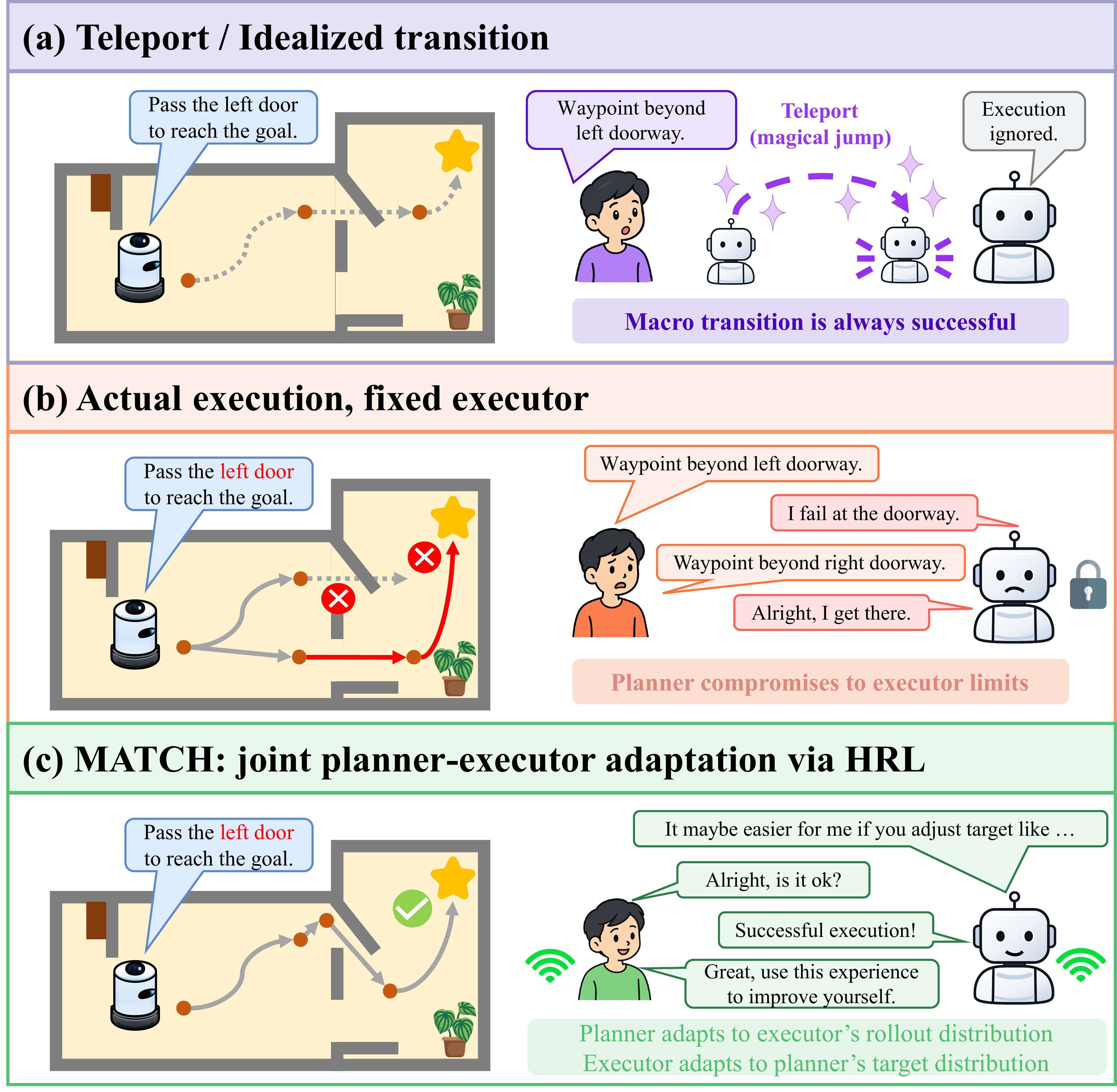}
    \caption{\small Comparison of three Hierarchical paradigms in VLN-CE.}
    \label{fig:teaser}
    \vspace{-20pt}
\end{figure}

Some methods bypass this execution challenge through teleportation~\cite{xue2026omninav, zhang2026navfom} (Fig.~\ref{fig:teaser}a), while some other methods rely on hand-designed controllers~\cite{an2024etpnav, hnr}.
Even when the planner is trained on actual execution outcomes~\cite{ye2026revisiting}, the execution policy remains fixed during that training.
The planner may learn to avoid a difficult doorway and select an alternative target, but the same experience does not teach how to handle that doorway more reliably (Fig.~\ref{fig:teaser}b).
Our insight is to extend the learning loop beyond macro-level planning to include micro-level execution: navigation experience should help to co-optimize the planner and executor, not just improve the planner's ability to work within its limits.

Hierarchical reinforcement learning (HRL)~\cite{sutton1999hrl} provides a natural framework for this joint adaptation. The planner determines the requests on which the executor learns, while the executor shapes the transitions those requests realize. The challenge is to couple these adaptations so that the executor learns on planner-generated requests and the planner selects requests that produce instruction-consistent progress under the current execution policy, rather than merely favoring easy targets. This calls for grounding macro-action learning both in the current executor and in the task contribution of its realized transitions.

We address this challenge with MATCH (\textbf{M}acro-\textbf{A}ction \textbf{T}ransition \textbf{C}oupling for \textbf{H}ierarchical RL) (Fig.~\ref{fig:teaser}c), which jointly learns a language-conditioned planner and a goal-conditioned executor.
We implement its decision interface within the graph-based VLN-CE architecture~\cite{an2024etpnav}, where waypoint proposals define candidate frontiers for an online topological planner.
To ground the planner's requests in the executor's evolving capabilities, we develop an \emph{execution-aware topological graph}.
In this graph, edges from different visited nodes to the same frontier represent distinct execution options, since execution outcomes depend not only on the destination but also on the starting location.
Each edge carries the executor's value estimates over a target neighborhood before execution and records the realized outcome memory afterward.
The planner pairs edge selection with a bounded continuous target offset within this evaluated neighborhood, allowing it to refine an execution request rather than merely accept or reject a fixed waypoint.

To learn which execution requests are useful for the task, we introduce a \emph{transition-aware dense macro reward} based on their realized transitions.
Each macro action receives credit for the ordered progress along the reference route, minus the execution residual---the remaining target-reaching distance---with both measured in meters.
The progress term encourages the planner to advance along the route, not just choose easier targets.
The residual gives the planner direct feedback on each request's execution difficulty, rather than leaving it implicit in final task outcomes.
This encourages executable requests without making local success a substitute for instruction following.
The executor is rewarded for progress toward assigned target.
A staged curriculum followed by synchronized on-policy updates lets both policies adapt through shared navigation experience.

Our main contributions are:
\begin{itemize}
    \item We propose MATCH, to our knowledge the first VLN-CE framework to jointly optimize a language-conditioned planner and a primitive-action executor through HRL, with both policies learning from shared on-policy navigation rollouts.
    \item To support joint adaptation during HRL, we develop an execution-aware topological graph that uses executor value estimates and outcome memory to guide the planner's selection and refinement of execution requests.
    \item We introduce a transition-aware dense macro reward that balances ordered progress along the reference route against execution residual, encouraging executable requests that advance the route.
    \item Using a 0.5B-parameter model, MATCH achieves state-of-the-art performance on R2R-CE and RxR-CE, with real-robot deployments further validating practicality. 
\end{itemize}

\input{MATCH_related_work}

\begin{figure*}[h]
    \centering
    \begin{minipage}[t]{1\linewidth}
    \centering
    \includegraphics[width=1\linewidth]{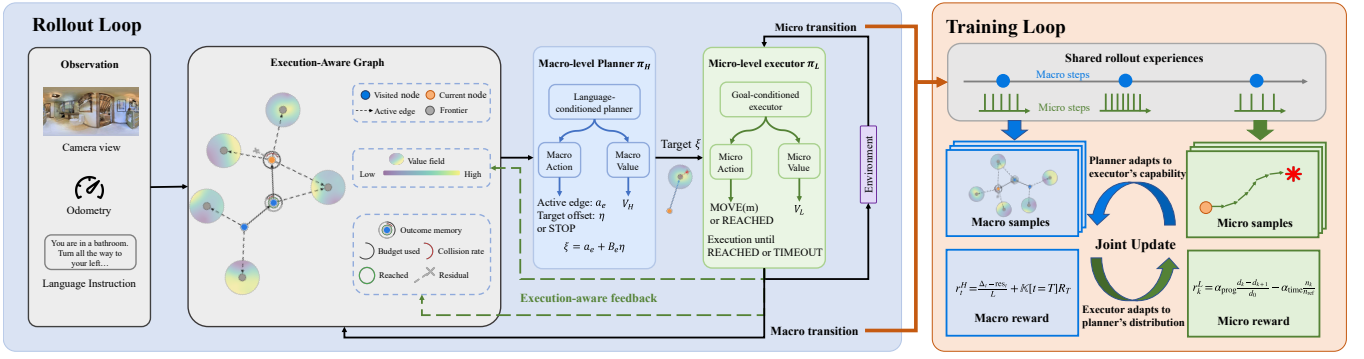}
    \end{minipage}
    \caption{\small{Overview of MATCH. The language-conditioned planner issues execution requests from an execution-aware graph, while the goal-conditioned executor realizes them with primitive actions. Macro- and micro-level experiences from the same rollouts are used for synchronized updates, enabling mutual adaptation between planning and execution.}}
    \label{fig:overview}
    \vspace{-15pt}
\end{figure*}

\section{Method}

\subsection{Hierarchical Formulation with a Learned Transition}\label{sec:formulation}

\textbf{Task.}
In VLN-CE an agent must follow a natural-language instruction $I$ to its goal in an unseen 3D environment.
At each step it observes a panoramic RGB-D image of 12 egocentric views and its odometry, and issues one primitive action: move forward by $\Delta s=0.25$\,m, turn left or right by $\Delta \theta$\, or stop.
It succeeds if it stops within 3\,m of the goal, typically after hundreds of primitives.

\textbf{Hierarchy.}
Following~\cite{ye2026revisiting}, we cast the task as a hierarchical MDP.
A waypoint predictor~\cite{hong2022bridging} predicts navigable waypoints around the agent, and the predicted waypoints are accumulated online into a topological graph $\mathcal G_t$~\cite{an2024etpnav}.
At macro step $t$ the planner acts on the macro state $s_t=(I,\mathcal G_t)$; its macro action commits the agent to one of the predicted waypoints or to STOP, and an executor $\pi_L$ realizes it with primitive actions before the graph is updated.
The macro transition $P^{\pi_L}(s_{t+1}\,|\,s_t,a_t)$ is thus determined by the environment together with the executor.
In~\cite{ye2026revisiting} the executor is a fixed heuristic, so this transition is fixed and only the planner learns.
MATCH learns both levels jointly as a hierarchical RL agent, so the transition itself changes during training, and the planner is adapted to this in three ways: it observes the executor's expected performance before committing to a target and the actual outcome afterwards, it can adjust the target within the neighborhood it has evaluated (Sec.~\ref{sec:graph}), and its reward is designed to account for both task progress and execution residual. (Sec.~\ref{sec:reward}).

\textbf{Workflow} (Fig.~\ref{fig:overview}).
At each macro step, the planner scores every active edge $e=(j,c)$ from a visited node $v_j$ to a frontier $c$, together with STOP, and samples an edge and a continuous offset $\eta$; together they define a 2D target $\xi$ near the location the edge points to.
If the agent does not stand at $v_j$, it first returns there by replaying the primitive sequences stored on the tree edges along the path; this needs no policy, so the executor is responsible only for reaching new targets.
The executor then drives toward $\xi$ until it declares REACHED or exhausts its decision budget (TIMEOUT).
Wherever it stops, a new visited node is created, the executed edge stores the outcome, waypoints predicted from the new node extend the graph, and every new edge is annotated with the executor's value estimates for targets around it.

\subsection{Goal-Conditioned Executor}\label{sec:executor}

\textbf{Interface.}
The executor receives no language; a sub-task is specified only by a 2D target $\xi$ relative to the start node.
The executor acts in \emph{decisions}, each of which expands to several primitives.
At decision $k$ it observes a single forward depth view, the distance $\rho_k$ and bearing $\theta_k$ of the target from odometry, and its previous decision.
A decision is either REACHED, which ends the sub-task, or $\mathrm{MOVE}(m)$ for one of $M$ ($M=360^\circ/\Delta \theta$) evenly spaced headings: the agent turns to heading $m$ with the fewest turn primitives and then moves forward once.
A sub-task also ends after $K_{\max}=16$ decisions (TIMEOUT).

\textbf{Reward.}
The executor is trained with privileged geometry that it never observes.
Let $\hat g$ be the navigable point closest to $\xi$; every target thus has a well-defined destination, as close to $\xi$ as the environment permits.
With $d_k$ the geodesic distance from the agent to $\hat g$ before decision $k$, a decision receives
\begin{equation}
r^L_k=\alpha_{\mathrm{prog}}\,\frac{d_k-d_{k+1}}{d_0}-\alpha_{\mathrm{time}}\,\frac{n_k}{n_{\mathrm{ref}}},\qquad n_{\mathrm{ref}}=\frac{d_0}{\Delta s},
\label{eq:lowreward}
\end{equation}
where $d_0$ is the initial distance, $n_k$ is the number of primitives the decision used.
Both terms are normalized by the sub-task size, so the executor's value $V_L$ is comparable across targets and indicates how reachable a target is expected to be, which Sec.~\ref{sec:graph} hands to the planner.

\subsection{Execution-Aware Topological Graph and Macro Actions}\label{sec:graph}

\textbf{Graph.}
$\mathcal G_t$ has three kinds of elements.
A \emph{visited node} $v_j$ is created at the start pose and at the end of every execution, wherever the agent stopped; it stores the agent's pose and the panorama observed there.
Every visited node except the root has exactly one parent, the node from which the agent reached it, so the visited nodes form a tree; the \emph{tree edge} from the parent stores the primitive sequence of that execution and its outcome memory, defined below.
A \emph{frontier} $c$ is a predicted waypoint that has not been visited yet; waypoints predicted from different visited nodes that lie within a merging radius of each other are merged into one frontier.
An \emph{active edge} $e=(j,c)$ connects a visited node $v_j$ to a frontier $c$ that was predicted from $v_j$.
Since a frontier merges the predictions of several nodes, the edge stores where $v_j$ itself predicted it: this position $a_e$ is the \emph{anchor} of the edge, and $\hat u_e$ is the unit direction from $v_j$ to $a_e$.
The graph is updated whenever a new visited node is created, by sorting the waypoints predicted from it into three cases.
A waypoint that lies close to a visited node is discarded.
A waypoint that lies close to an existing frontier is merged into it, and a new active edge connects the new node to that frontier.
Any other waypoint becomes a new frontier, connected to the new node in the same way.
The executed edge is removed, and if the agent reached its frontier, the frontier is removed as well.
Edges are kept specific to their parent node because the outcome of an execution depends not only on the target but also on where it starts.

\textbf{Macro action.}
A macro action is either STOP or an active edge together with a target offset:
\begin{equation}
\pi_H(e,\eta\,|\,s_t)=\pi_E(e\,|\,s_t)\;\pi_{\mathrm{off}}(\eta\,|\,e,s_t),\qquad \xi=a_e+B_e\,\eta,
\label{eq:macro}
\end{equation}
where $B_e=[\hat u_e,\hat u_e^{\perp}]$ is the frame of the edge and $\eta$ lies in the disk $\mathbb D$ of radius $r_{\max}=1$\,m around the anchor.
$\pi_E$ is a categorical distribution over the active edges and STOP, and $\pi_{\mathrm{off}}$ is a Gaussian squashed radially onto $\mathbb D$.

\textbf{Value field: what the planner sees before execution.}
When an active edge is created, it is annotated with the executor's own value estimates for targets around its anchor:
\begin{equation}
\mathcal F_e(\eta_n)=V_L\big(s^0_e,\ a_e+B_e\eta_n\big),\qquad n=1,\dots,N_F,
\label{eq:field}
\end{equation}
where $s^0_e$ is the executor's initial state for this edge, and the $N_F=16$ offsets $\eta_n$ form a uniform distribution over $\mathbb D$, with $\eta_1=0$ at the anchor.
$V_L$ estimates the normalized return for each target under the current execution policy, providing policy-dependent feedback for the planner’s selection and refinement of execution requests.
$\mathcal F_e$ enters the active edge token of the planner network.

\textbf{Outcome memory: what the planner sees after execution.}
After execution along active edge $e$ with target $\xi$, the newly created edge between the last and current node records
\begin{equation}
H_e=\big[\,\mathbb 1_{\mathrm R},\ k_{\mathrm{used}}/K_{\max},\ n^{\mathrm{coll}}/n^{\mathrm{fwd}},\ B_e^{\top}(\xi-p^{\mathrm{end}})/\rho_0\,\big]
\label{eq:memory}
\end{equation}
Its entries are, in order, whether the executor declared REACHED, the fraction of the decision budget used, the fraction of forward primitives that collided, and the horizontal displacement from the final position $p^{\mathrm{end}}$ to the target, expressed in the edge frame and normalized by the initial horizontal distance $\rho_0$.
$H_e$ is attached to the child node and enters its token of the planner network.

\subsection{Transition-Aware Macro Reward}\label{sec:reward}

To help the planner accurately assess the difficulty of the requests it issues to the executor, we incorporate the execution residual of each request into the macro reward. However, it may also encourage the planner to generate overly simple requests, disregarding task progress. A complementary reward that measures task progress is therefore necessary to balance this tendency. How to quantify progress at each macro step is thus an important engineering problem. We first introduce our solution, ordered route coverage, and then describe how it is integrated into the macro reward.

\textbf{Ordered route coverage.}
The progress should measure motion along the reference route.
Alignment-based metrics such as nDTW~\cite{ilharco2019ndtw} may appear to be a natural choice. However, they measure how closely the executed trajectory aligns with the reference route as a whole, rather than how much progress has been made along that route.

We instead propose ordered route coverage.
Let the reference route be a polyline of length $L$, resampled at spacing $\Delta s$ into $K$ cells with centers $\gamma_\nu$ and unit tangents $\tau_\nu$.
The trajectory so far, replays included, is a sequence of forward primitives $i=1,\dots,n$, each ending at position $x_i$ with heading $u_i$.
Forward primitive $i$ may cover cell $\nu$ with weight
\begin{equation}
\kappa_{i\nu}=(1-c_i)\kappa\big(\|x_i-\gamma_\nu\|\big)\cdot\mathbb 1\big[u_i^{\top}\tau_\nu\ge0\big],
\label{eq:kappa}
\end{equation}
where $c_i$ indicates that a collision occurs during the \(i\)-th forward movement, $\mathbb 1[\cdot]$ is the indicator function, and $\kappa(d)$ equals 1 for $d\le0.5$\,m and decays linearly to 0 at $d=2.5$\,m, so that walking beside the route still counts, while motion against the direction of the route does not.
The coverage $A_t$ is the total length of the cells that can be matched to forward primitives one-to-one and in order.
The maximum-weight matching is found by dynamic programming,
\begin{equation}
G_{i\nu}=\max\{\,G_{i-1,\nu},\ G_{i,\nu-1},\ G_{i-1,\nu-1}+\kappa_{i\nu}\Delta s\,\},
\label{eq:dp}
\end{equation}
with $G_{0,\cdot}=G_{\cdot,0}=0$ and $A_t=G_{n_t,K}$, $n_t$ being the number of forward primitives executed before macro step $t$; the three terms of the maximum correspond, in order, to skipping primitive $i$, to skipping cell $\nu$, and to matching the two.
The coverage progress of macro step $t$ is $\Delta_t=A_{t+1}-A_t$.
Since $A_t$ can only grow, $\Delta_t\ge0$; it is a length in meters that never exceeds the distance walked in the step.

\textbf{Reward.}
The planner receives reward at macro step $t$
\begin{equation}
r^H_t=\frac{\Delta_t-\mathrm{res}_t}{L}+\mathbb 1[t=T]\,R_T,
\label{eq:reward}
\end{equation}
where the residual $\mathrm{res}_t=D_{\mathrm{geo}}(p^{\mathrm{end}}_t,\hat g_t)$ is the geodesic distance from the agent's final position to the reachable point $\hat g_t$ of the assigned target, $R_T=\mathrm{SR}+\mathrm{SPL}+\mathrm{nDTW}$ is the terminal reward, and a STOP step has $\Delta_t=\mathrm{res}_t=0$.
The two dense terms pull in opposite directions and balance each other.
The coverage term rewards targets far along the route, the residual penalizes targets the executor could not reach.
Because both are in meters, one meter of route covered is worth exactly one meter of target left unreached: a far target pays off only if the executor is expected to get there, and a very close target is always reached but earns little coverage, so the planner is pushed neither to overreach nor to collapse to simple targets.
The residual is charged even when the miss is due to the executor's own randomness, since the planner is responsible for how its choice plays out and the value field lets it anticipate it; one navigation experience thus trains both levels, through Eq.~\eqref{eq:lowreward} and Eq.~\eqref{eq:reward}.

\subsection{Network Architecture}\label{sec:network}
While the interfaces above fix the inputs and outputs of both levels we leave the details of network architecture to \hyperref[app:network]{the appendix}.

\section{Training}\label{sec:training}

Training the two levels of MATCH jointly from the start is not practical.
With an untrained executor, the value fields carry no information and the planner's targets are never reached; with an untrained planner, the executor is requested to arbitrary places.
Neither level can learn from the other until both are competent, so we train in three stages as below.

\subsection{Stage 1: Parallel Warmup}

\textbf{Planner.}
The planner uses the same data and methods as Revisiting~\cite{ye2026revisiting} for offline pretraining.
No executor is involved at this stage, so the value fields and outcome memories of all edges and nodes are filled with a learned placeholder, and the offset head and the value head are not trained.
What the planner learns here is grounding: how the instruction and the panoramas relate to the choice of a frontier.

\textbf{Executor.}
The executor is pretrained on sub-tasks generated offline.
A sub-task consists of a start node sampled from the expert trajectories and a 2D target generated by the waypoint predictor~\cite{hong2022bridging}.
Supervision comes from the simulator's oracle shortest path follower with access to the navigable mesh.
The executor is first trained by behavior cloning on the teacher's decisions, then by DAgger, and finally by PPO~\cite{PPO} with the reward of Eq.~\eqref{eq:lowreward}.

\subsection{Stage 2: Online Imitation with a Frozen Executor}\label{sec:stage2}

The planner is now trained by DAgger with the same teacher supervision as in~\cite{an2024etpnav}.
The frozen executor of stage 1 realizes its macro actions, so the planner faces the states a real executor produces and receives value fields and outcome memories as inputs.
These inputs are not yet clearly related to the decisions: the teacher has no mechanism to reflect the executor's reachability in its labels, so imitation can only accustom the planner to the form of the execution feedback, while learning to use it is left to the stage 3.
For the same reason the offset head is not trained here and starts stage 3 from a zero offset, by initializing the last layer of the offset mean head to zero.
After imitation, $V_H$ is fitted to the returns of Eq.~\eqref{eq:reward} on autonomous rollouts of the frozen policy.

\subsection{Stage 3: Joint Hierarchical Reinforcement Learning}

Both levels are now trained by PPO on the same experience.
Each iteration collects a batch of full episodes with both policies sampling, uses it for one epoch of updates of both levels, and discards it.
The episodes of this stage are held out from stage 2, so that the planner does not re-encounter episodes it imitated: 80\% of the training episodes are used for stage 2 and the remaining 20\% for stage 3.

\textbf{Planner update.}
The planner is updated on macro steps of the batch, with the importance ratio of the joint macro-action probability $\pi_E(e\,|\,s_t)\,\pi_{\mathrm{off}}(\eta\,|\,e,s_t)$ and advantages estimated by GAE over macro steps from the reward of Eq.~\eqref{eq:reward} with $\gamma_H=1$.
The loss is the clipped PPO surrogate with a value loss on $V_H$ and an entropy bonus on $\pi_E$ and $\pi_{\mathrm{off}}$.

\textbf{Executor update.}
The executor is updated on the decisions of all sub-tasks in the batch, with the importance ratio of the decision taken under $\pi_L$ and advantages estimated by GAE within each sub-task from the reward of Eq.~\eqref{eq:lowreward} with $\gamma_L=1$.
The loss is the clipped PPO surrogate with a value loss on $V_L$ and an entropy bonus. Thus, executor training follows the starting-state and target distribution encountered under the current hierarchy, rather than a separate, fixed distribution of execution sub-tasks.

\section{Experiments}\label{sec:exp}

\subsection{Experimental Setup}

\textbf{Benchmarks and metrics.}
We evaluate on the val-unseen splits of R2R-CE~\cite{VLN} and RxR-CE~\cite{RXR} in habitat simulator~\cite{savva2019habitat}.
R2R-CE provides English instructions for shortest-path routes.
RxR-CE provides multilingual instructions for routes that are substantially longer and possibly non-shortest paths.
The two benchmarks also differ in their agents: R2R-CE has $\Delta \theta=15^\circ$ and allows sliding along obstacles, whereas RxR-CE's $\Delta \theta=30^\circ$, has a larger agent radius and forbids sliding.
We report in the standard navigation metrics: navigation error (NE), oracle success rate (OS), success rate (SR) and success weighted by path length (SPL) and normalized dynamic time warping (nDTW).

\textbf{Implementation details.}
Architectural details are given in \hyperref[app:network]{the appendix}.
$\alpha_{\mathrm{prog}}$ and $\alpha_{\mathrm{time}}$ in Eq.~\eqref{eq:lowreward} are 1.0 and 0.1 respectively.
For Training Stage 3, Each iteration of the joint stage collects 128 episodes with the current policies and performs one update of both levels. We totally run 4000 iterations.
PPO uses a clipping range of 0.2, GAE with $\lambda=0.95$, a value-loss coefficient of 0.5 and an entropy coefficient of 0.01.
Training runs on 4 NVIDIA RTX 5880 GPUs and takes about 70\,h for the planner and 100\,h for the executor in stage 1, 90\,h in stage 2 and 120\,h in the joint stage.

\begin{table*}[t]
\caption{Comparison on the val-unseen splits of R2R-CE and RxR-CE. $\ddagger$: reproduced with the released checkpoint the rotate-then-forward executor, since the original evaluation follows the predicted waypoint trajectory by teleportation.}
\label{tab:main}
\centering
\footnotesize
\setlength{\tabcolsep}{2.3pt}
\begin{tabular}{l c c l l l cccc cccc}
\toprule
\multirow{2}{*}{Method} & \multirow{2}{*}{Params} & \multirow{2}{*}{Training} & \multirow{2}{*}{Planner output} & \multirow{2}{*}{Executor} & \multirow{2}{*}{Observation} & \multicolumn{4}{c}{R2R-CE Val-Unseen} & \multicolumn{4}{c}{RxR-CE Val-Unseen}\\
\cmidrule(lr){7-10}\cmidrule(lr){11-14}
 & & & & & & NE$\downarrow$ & OS$\uparrow$ & SR$\uparrow$ & SPL$\uparrow$ & NE$\downarrow$ & SR$\uparrow$ & SPL$\uparrow$ & nDTW$\uparrow$\\
\midrule
ETPNav~\cite{an2024etpnav} & 0.3B & IL & frontier & rotate-then-fwd. & pano.\,RGB-D+odo. & 4.71 & 65.0 & 57.0 & 49.0 & 5.64 & 54.8 & 44.9 & 61.9\\
HNR~\cite{hnr} & 0.3B & IL & frontier & rotate-then-fwd. & pano.\,RGB-D+odo. & 4.42 & 67.0 & 61.0 & 51.0 & 5.51 & 56.4 & 46.7 & 63.6\\
Revisiting~\cite{ye2026revisiting} & 0.5B & IL+RL & frontier & rotate-then-fwd. & pano.\,RGB-D+odo. & 3.89 & 72.6 & 68.1 & 57.3 & 5.07 & 62.3 & 50.2 & 66.8\\
\midrule
VLN-R1~\cite{qi2025vlnr1} & 7B & IL+RL & primitives & -- & mono RGB & 7.00 & 41.2 & 30.2 & 21.8 & 10.4 & 22.3 & 17.5 & --\\
StreamVLN~\cite{wei2025streamvln} & 7B & IL & primitives & -- & mono RGB & 4.98 & 64.2 & 56.9 & 51.9 & 6.22 & 52.9 & 46.0 & 61.9\\
DualVLN~\cite{wei2026dualvln} & 7B & IL & pixel goal & learned policy & mono RGB & 4.05 & 70.7 & 64.3 & 58.5 & 4.58 & 61.4 & 51.8 & \textbf{70.0}\\
OmniNav$^\ddagger$~\cite{xue2026omninav} & 3B & IL & trajectory & rotate-then-fwd. & 3-view RGB & 4.06 & 73.1 & 66.1 & 61.1 & 4.75 & 64.7 & 53.0 & 69.6\\
\midrule
MATCH & 0.5B & IL+HRL & edge + offset & learned policy & pano.\,RGB-D+odo. & \textbf{3.78} & \textbf{73.3} & \textbf{69.0} & \textbf{61.5} & \textbf{4.51} & \textbf{67.7} & \textbf{54.2} & \textbf{70.0}\\
\bottomrule
\vspace{-20pt}
\end{tabular}
\end{table*}

\subsection{Comparison with the State of the Art}
Table~\ref{tab:main} compares MATCH with previous methods on both benchmarks.

The first group follows the same high-level architecture as MATCH: a planner selects frontiers on a topological graph, while low-level execution is handled by a heuristic controller.
Revisiting is the strongest of them but MATCH improves over it on every metric, by 0.9 SR and 4.2 SPL on R2R-CE and by 5.4 SR and 3.2 nDTW on RxR-CE, at a comparable model size.

The second group contains recent methods built on large vision-language models.
VLN-R1 and StreamVLN output primitive actions directly from the language model.
DualVLN is a dual-system method whose VLM grounds a pixel goal in the current image and a diffusion policy executes it.
It is therefore the closest in spirit to MATCH, but its two systems are trained sequentially and the high-level model receives no feedback from the low-level one.
MATCH surpasses DualVLN by 4.7 SR and 3.0 SPL on R2R-CE and by 6.3 SR and 2.4 SPL on RxR-CE, with about 1/10 of its parameters.
OmniNav reports its results with a teleporting executor that follows the predicted trajectory waypoints without executing primitive actions.
To compare it under the same conditions as every other method in the table, we reproduce its evaluation with the rotate-then-forward executor.
Under this setting MATCH surpasses it by 2.9 SR in R2R-CE and 3.0 SR in RxR-CE.

\begin{table}[t]
\caption{Ablations on RxR-CE val-unseen. $\Delta$ is relative to MATCH; time is the wall-clock of the joint stage in hours.}
\label{tab:ablation}
\centering
\footnotesize
\setlength{\tabcolsep}{3.5pt}
\begin{tabular}{l cc cc c}
\toprule
Variant & SR$\uparrow$ & nDTW$\uparrow$ & $\Delta$SR & $\Delta$nDTW & Time\\
\midrule
MATCH & 67.7 & 70.0 & -- & -- & $\sim$120\\

\midrule
\multicolumn{6}{l}{\textit{Training (Sec.~\ref{sec:executor}, \ref{sec:training})}}\\
w/o joint HRL & 61.8 & 65.1 & $-5.9$ & $-4.9$ & --\\
heuristic executor & 64.3 & 66.9 & $-3.4$ & $-3.1$ & $\sim$50\\
learned executor, frozen & 65.6 & 68.0 & $-2.1$ & $-2.0$ & $\sim$110\\
alternating updates & 67.0 & 69.8 & $-0.7$ & $-0.2$ & $\sim$200\\
\midrule
\multicolumn{6}{l}{\textit{Execution-aware graph (Sec.~\ref{sec:graph})}}\\
w/o value field & 66.1 & 68.2 & $-1.6$ & $-1.8$ & $\sim$120\\
w/o outcome memory & 65.5 & 69.0 & $-2.2$ & $-1.0$ & $\sim$120\\
one edge per frontier & 66.2 & 69.3 & $-1.5$ & $-0.7$ & $\sim$120\\
\midrule
\multicolumn{6}{l}{\textit{Adjusting the target (Sec.~\ref{sec:graph})}}\\
no offset ($\eta=0$) & 64.8 & 67.9 & $-2.9$ & $-2.1$ & $\sim$120\\
\midrule
\multicolumn{6}{l}{\textit{Reward design (Sec.~\ref{sec:reward})}}\\
terminal reward only & 65.1 & 68.0 & $-2.6$ & $-2.0$ & $\sim$120\\
w/o residual term & 65.5 & 68.1 & $-2.2$ & $-1.9$ & $\sim$120\\
w/o coverage term & 62.2 & 65.5 & $-5.5$ & $-4.5$ & $\sim$120\\
nDTW increment as progress & 66.2 & 69.1 & $-1.5$ & $-0.9$ & $\sim$120\\
\bottomrule
\vspace{-20pt}
\end{tabular}
\end{table}

\subsection{Ablation Study}
To validate our proposed framework, we conduct ablation studies on the RxR-CE Val-Unseen split. Table~\ref{tab:ablation} summarizes the results and training cost.

\textbf{Training.}
The result of ``w/o joint HRL'' first shows the significant total performance gains in RxR-CE of training stage 3 (5.9 SR and 4.9 nDTW). 
Replacing the learned executor by the heuristic rotate-then-forward executor, while training the planner with the same curriculum, costs 3.4 SR and 3.1 nDTW.
In this variant the value field are replaced by a learned placeholder, since a heuristic executor has no value function, so the row measures the executor together with the value feedback that only a learned executor can provide.
Keeping the learned executor frozen after Stage 1 reduces SR by 2.1 and nDTW by 2.0. This supports the task-level benefit of continuing executor training on the requests generated by the evolving planner, rather than adapting the planner alone to a fixed execution policy.
Finally, updating the two levels alternately, one level per round with the other frozen, reaches almost the same accuracy as synchronous updates.
But it takes 200\,h instead of 120\,h, because the episodes collected while one level is trained are not used to train the other.
Synchronous training is thus the more economical way of training both levels from the same experience.

\textbf{Execution-aware graph.}
Removing the value field costs 1.6 SR and 1.8 nDTW, and removing the outcome memory costs 2.2 SR and 1.0 nDTW.
The two channels play different roles.
The value field acts before the planner's decision: it helps the planner choose targets that the executor can reach and that advance along the route, which shows more in route fidelity.
The outcome memory acts after the execution: it tells the planner where an execution ended and whether it failed, which mainly keeps the planner from repeating an unreachable target and shows more in success.
Restricting every frontier to a single edge from its nearest parent node costs 1.5 SR and 0.7 nDTW. This implies that disregarding the starting point of the frontier can influence the planner's assessment of low-level execution capabilities.

\textbf{Adjusting the target.}
Fixing the target to the anchor, so that the planner can only choose which edge to execute, is the most costly single change among the graph ablations (2.9 SR and 2.1 nDTW).
This highlights the improvements in instruction following and reachability of the selected requests achieved through offset adjustment.

\textbf{Reward design.}
Training the planner with the terminal reward alone, costs 2.6 SR and 2.0 nDTW.
Adding the coverage term without the residual recovers only 0.4 SR of this loss.
Because this is merely an approximation of distributing the terminal reward across each step.
Keeping the residual without the coverage term is the worst variant of all, 5.5 SR and 4.5 nDTW below MATCH.
This is the collapse anticipated in Sec.~\ref{sec:reward}: a planner that is charged for missed targets learns to assign simple targets that are always reached, but too slowly to finish the episodes.
The two terms therefore need each other, and their balance is what makes the dense reward useful.
Finally, replacing ordered route coverage by the per-step increment of nDTW, the most natural alternative for a progress measure, costs 1.5 SR and 0.9 nDTW.
This is because nDTW only measures the similarity between two trajectories, whereas the ordered route coverage we defined is explicitly designed to incorporate incremental task progress.

\begin{figure}[t]
\centering
\includegraphics[width=\columnwidth]{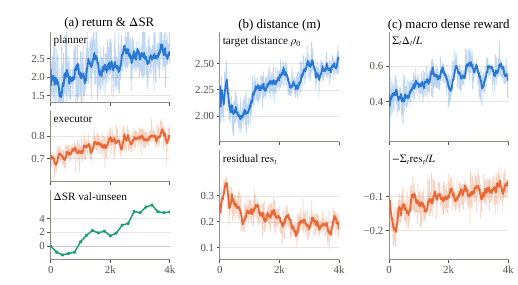}
\caption{Training dynamics during Stage~3 joint HRL. The x-axis in all panels denotes the training step. (a) Planner and executor returns, together with the change in success rate ($\Delta\mathrm{SR}$) on RxR-CE val-unseen. (b) Planner-proposed target distance $\rho_0$ and the execution residual $\mathrm{res}_t$ after each macro action. (c) The two dense components of the macro reward: normalized ordered-route coverage increment reward, $\sum_t \Delta_t/L$, and normalized execution-residual penalty, $-\sum_t \mathrm{res}_t/L$. Together, these curves illustrate how the planner gradually balances target ambition with executability as the planner and executor co-adapt.}
\label{fig:curves}
\vspace{-20pt}
\end{figure}

\begin{figure*}[t]
\centering
\includegraphics[width=1\linewidth]{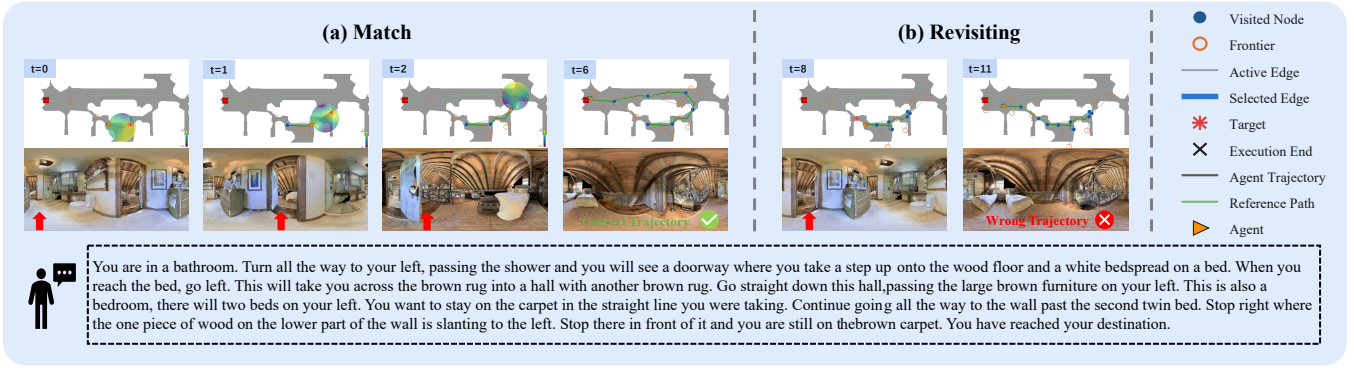}
\caption{Comparative case study on RxR-CE val-unseen.}
\label{fig:case}
\end{figure*}

\begin{figure*}[t]
\centering
\includegraphics[width=1\linewidth]{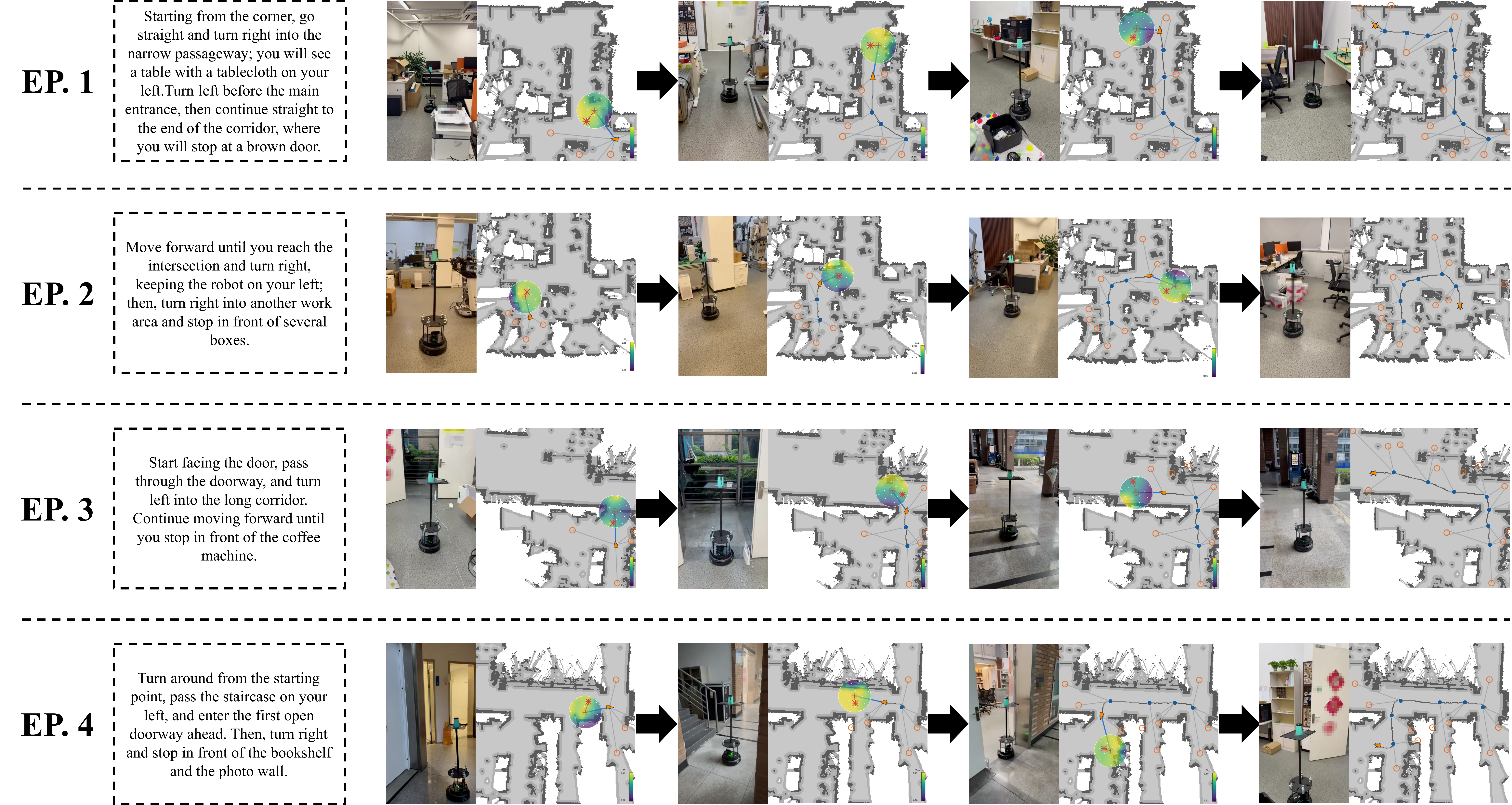}
\caption{Real robot deployment. We deployed MATCH on a robot modified from TurtleBot 4 and tested it across four episodes.}
\label{fig:real}
\vspace{-20pt}
\end{figure*}

\subsection{Training Dynamics and Qualitative Analysis}
Fig.~\ref{fig:curves} shows Stage~3 training dynamics. Early in training, the
planner increases target distance before the offset policy is well
trained, which increases the execution residual. It then shortens
targets and gradually increases their distance again as joint
training stabilizes.

Fig.~\ref{fig:case} shows one representative episode in RxR-CE val-unseen.
In this episode, the robot is required to navigate through two consecutive narrow passages; however, the environment also features a slightly wider alternative route that leads to the target but contradicts the instructions. Revisiting~\cite{ye2026revisiting} initially attempts the instructed path but fails to pass through, subsequently opting to reach the destination via the incorrect route. In contrast, Match leverages the cooperation between its planner and executor, successfully navigates the narrow passages and reaches the destination by following the correct trajectory.

\subsection{Real-World Deployment}
We deploy MATCH on a TurtleBot 4 equipped with an elevated Seeker multi-camera system that provides four-view panoramic RGB observations. An onboard NVIDIA Jetson Orin interfaces with the camera system and handles image processing and transmission, while model inference is performed on a remote server equipped with an NVIDIA RTX 3090 GPU. All system components communicate through ROS~\cite{macenski2022ros}. Since MATCH additionally requires panoramic depth and robot pose as inputs, we use FoundationStereo~\cite{wen2025foundationstereo} to estimate depth for the four camera views. For localization, we first construct a map of the environment using SLAM~\cite{macenski2021slam} and then estimate the robot pose online using Adaptive Monte Carlo Localization (AMCL)~\cite{fox1999amcl}. Importantly, the pre-built map is used only for localization and visualization and is never provided to MATCH.

The physical sensing configuration differs from the original MATCH setup, which represents each graph node using 12 visual tokens. We therefore first retrain the waypoint predictor using four-view observations. For the planner, we introduce a lightweight attention-based adapter that lifts the four tokens associated with each node to the original 12-token representation, while leaving the remainder of the planner architecture unchanged. We train this adapter and the modified planner by distilling knowledge from the original MATCH planner, followed by Stage 3 joint training in simulation using the same four-view camera configuration as the real robot.

During real-world deployment, the planner operates at approximately 0.1 Hz, whereas the executor runs at approximately 1 Hz. We evaluate the complete system over four episodes in indoor environments, with results summarized in Fig.~\ref{fig:real}. The results indicate that MATCH transfers effectively from simulation to the physical platform and exhibits promising sim-to-real generalization.

\section{Conclusion}

We presented MATCH, a hierarchical reinforcement learning framework that improves VLN-CE navigation through joint adaptation of a language-conditioned planner and a goal-conditioned executor. Its execution-aware graph, continuous target refinement, and transition-aware reward connect the executor’s evolving capabilities to progress along the instructed route. Ablation studies support the benefits of joint adaptation and execution feedback, while highlighting the importance of balancing route progress with execution residual. Real-world deployment across four indoor real-robot episodes also suggests promising sim-to-real transfer. Together, these findings support learning what to execute and how to execute it through shared experience, rather than adapting planning alone to a fixed controller.

\putbib[root]
\end{bibunit}

\clearpage
\begingroup
\normalfont\normalsize
\setlength{\abovedisplayskip}{1.5ex plus 4pt minus 2pt}
\setlength{\belowdisplayskip}{1.5ex plus 4pt minus 2pt}
\setlength{\abovedisplayshortskip}{0pt plus 4pt}
\setlength{\belowdisplayshortskip}{1.5ex plus 4pt minus 2pt}
\makeatletter
\let\maketitle\MATCHmaketitle
\let\@maketitle\MATCHatmaketitle
\let\caption\MATCHclasscaption
\let\@caption\MATCHclassatcaption
\let\@makecaption\MATCHclassmakecaption
\abovecaptionskip=\MATCHabovecaptionskip
\belowcaptionskip=\MATCHbelowcaptionskip
\def\@extra@b@citeb{.appendix}
\makeatother
\IEEEaftertitletext{}
\title{\LARGE \bf Appendix: Network Architecture of MATCH}
\makeatletter
\let\@author\MATCHclassauthor
\makeatother
\setcounter{section}{0}
\setcounter{subsection}{0}
\setcounter{subsubsection}{0}
\setcounter{paragraph}{0}
\setcounter{equation}{0}
\setcounter{figure}{0}
\setcounter{table}{0}
\renewcommand{\theHsection}{app.\arabic{section}}
\renewcommand{\theHequation}{app.\arabic{equation}}
\renewcommand{\theHfigure}{app.\arabic{figure}}
\renewcommand{\theHtable}{app.\arabic{table}}
\hypersetup{hidelinks}
\begin{bibunit}[ieeetr]
\def\bibliography#1{\putbib[#1]}
\makeatletter
\let\bibliographystyle\@gobble
\makeatother
\input{appendix}

\clearpage
\end{bibunit}
\endgroup

\end{document}

%% file: MATCH_related_work.tex

\section{Related Work}
\label{sec:related_work}

\subsection{Hierarchical Vision-and-Language Navigation}
VLN-CE~\cite{krantz2020beyond} requires instruction following through collision-constrained primitive actions rather than assuming oracle transitions between predefined viewpoints.
Waypoint prediction~\cite{hong2022bridging} introduces intermediate spatial goals to connect language-conditioned planning with movement, enabling discrete-environment agents to transfer to continuous settings without predefined navigation graphs.
Graph-based methods extend planning context: DUET~\cite{chen2022duet} combines fine-grained local grounding with global topological reasoning, while BEVBert~\cite{an2022bevbert} integrates metric and topological maps for spatially grounded navigation.
ETPNav~\cite{an2024etpnav} couples online topological mapping and cross-modal planning with a heuristic controller that tries alternative actions when execution is blocked.
DualVLN~\cite{wei2026dualvln} learns a diffusion-based executor conditioned on a supervised pixel-goal planner, but does not refine the planner through RL on executor-induced macro transitions.
HSAN~\cite{fang2026hsan} combines supervised high-level planning with RL-based low-level control.
Ye et al.~\cite{ye2026revisiting} optimize topological frontier selection through closed-loop macro-action RL while retaining a fixed controller.
MATCH differs by applying RL to both levels, jointly optimizing a language-conditioned planner and a primitive-action executor from shared on-policy navigation rollouts.

\subsection{Hierarchical Learning and Execution Feedback}

In state-based control, goal-conditioned HRL methods jointly learn high-level subgoals and low-level policies~\cite{zhang2020hrac, lee2022dhrl}.
HIRO~\cite{nachum2018hiro} corrects off-policy experience to account for changing low-level behavior, while HAC~\cite{levy2017hac} uses hindsight transitions to learn multiple policy levels in parallel.
For execution grounding, SayCan~\cite{ahn2022saycan} uses value functions of pretrained skills to inform language-level decisions.
For route-aware supervision and evaluation, LAW~\cite{raychaudhuri2021law} aligns corrective actions with the instructed path instead of only the destination, while nDTW~\cite{ilharco2019ndtw} measures ordered path fidelity and can serve as an RL reward.

MATCH couples macro-action learning to both the executor's evolving capabilities and the task contribution of realized transitions.
Its execution-aware topological graph uses current executor values and outcome memory to evaluate source--target execution options, guiding request selection and bounded continuous target refinement.
Its transition-aware macro reward credits ordered route progress while penalizing execution residual, providing direct feedback on both task contribution and execution shortfall.
Together, these mechanisms support joint adaptation without treating local target completion as the navigation objective.

%% file: appendix.tex
\maketitle
\thispagestyle{empty}
\pagestyle{empty}

\section{Network Architecture}\label{app:network}

This appendix specifies the networks that implement the two levels of MATCH.
Their interfaces are fixed in the main paper: the planner maps the instruction and the execution-aware graph $\mathcal G_t$ to a distribution $\pi_E$ over the active edges and STOP, an offset distribution $\pi_{\mathrm{off}}$ for the selected edge, and a value $V_H$; the executor maps its observation and the relative target to a distribution over decisions and a value $V_L$, and this value, evaluated at the start of a sub-task for targets around an anchor, is the value field of an edge.
Sections~\ref{app:encoders}--\ref{app:heads} describe the planner from the encoders to the heads, Sec.~\ref{app:executor} the executor, and Table~\ref{tab:app-modules} lists all modules with their sizes, initialization and training status in each stage.
Throughout, $d=768$ denotes the hidden size of the planner and $\mathrm{MLP}$ a two-layer perceptron with a ReLU in between.


\subsection{Encoders}\label{app:encoders}

\textbf{RGB.}
Each of the 12 views of a panorama is encoded by the SigLIP2 ViT-B/16 image encoder~\cite{tschannen2025siglip2} at a resolution of $224\times224$; its pooled embedding is projected linearly to $d$.
The encoder is frozen in all stages.

\textbf{Depth.}
Depth views, at $256\times256$, are encoded by the ResNet-50 depth encoder of the DD-PPO PointNav agent~\cite{wijmans2019ddppo}, including the output projection of that agent, which maps the ResNet features to a 512-dimensional vector; the planner projects this vector linearly to $d$.
The same frozen encoder serves the executor (Sec.~\ref{app:executor}), so a depth view is encoded once and used by both levels.

\textbf{Instruction.}
The instruction is encoded by InfoXLM-base~\cite{chi2021infoxlm}, a 12-layer multilingual Transformer with hidden size 768 whose output tokens are used directly as the instruction tokens of the planner.
InfoXLM is fine-tuned in every stage in which the planner is trained.

\textbf{Panorama.}
For each view $v$ of a visited node $j$, the RGB feature, the depth feature and a sinusoidal encoding of the view's heading relative to the node's heading are concatenated and projected to $d$.
The 12 vectors then pass through two Transformer encoder layers with self-attention over the views (12 heads, feed-forward size $4d$).
The outputs are the view tokens $\tilde{\mathbf z}^{(j)}_v$, $v=1,\dots,12$.

\subsection{Planner Input Tokens}\label{app:tokens}

The planner's input sequence consists of the instruction tokens and three kinds of graph tokens: one token per visited node, one per active edge, and one STOP token.
Graph tokens carry only content, that is, what a place looks like, what the executor expects there and how executions there went.
A learned type embedding $\mathbf e_{\mathrm{node}}$, $\mathbf e_{\mathrm{edge}}$ or $\mathbf e_{\mathrm{stop}}$ is added to every graph token.

\textbf{Node token.}
The token of visited node $j$ is
\begin{equation}
\mathbf h_j=\mathrm{AttnPool}\big(\{\tilde{\mathbf z}^{(j)}_v\}_{v=1}^{12}\big)+\mathrm{MLP}(H_j)+\mathbf e_{\mathrm{node}},
\label{eq:app-node}
\end{equation}
where $\mathrm{AttnPool}$ is attention pooling with a single learned query, which summarizes the panorama into one vector, and $H_j\in\mathbb R^5$ is the outcome memory of the execution that created the node.
The root node has no such execution and uses a learned placeholder, the ``missing'' vector below, in place of $\mathrm{MLP}(H_j)$.

\textbf{Edge token.}
The token of an active edge $e=(j,c)$ with anchor $a_e$ describes the sub-task that the edge represents from the executor's point of view:
\begin{equation}
\mathbf h_e=\mathrm{LN}\big(W_1\tilde{\mathbf z}^{(j)}_{v(e)}+W_2\bar{\mathbf z}_c+W_3\bar{\mathbf f}_e\big)+\mathbf e_{\mathrm{edge}}.
\label{eq:app-edge}
\end{equation}
The four terms are the following.
$\tilde{\mathbf z}^{(j)}_{v(e)}$ is the view token of the parent node whose heading is closest to the direction $\hat u_e$ of the anchor; it is what the executor will face when it starts the sub-task.
$\bar{\mathbf z}_c$ is the mean of these anchor-facing view tokens over all active edges that lead to frontier $c$; it describes the frontier itself and is shared by all edges that lead to it, which tells the planner that edges from different parents point to the same place.
$\bar{\mathbf f}_e=\mathrm{MLP}\big([\mathcal F_e(\eta_1),\dots,\mathcal F_e(\eta_{N_F})]\big)$ encodes the value field of the edge; since the $N_F=16$ samples lie on a fixed uniform distribution, their order in the vector identifies their positions, so no positional information needs to be added.
$W_1,\dots,W_3$ are linear maps to $d$ and $\mathrm{LN}$ is layer normalization.

\textbf{STOP token.}
The STOP token is a learned vector plus an $\mathrm{MLP}$ encoding of the episode's progress, the fractions of the macro-step budget and of the primitive budget that have been used.

\textbf{Missing vectors.}
When a value field is unavailable, its sampled value vector $[\mathcal F_e(\eta_1),\dots,\mathcal F_e(\eta_{N_F})]$ is replaced by a learned placeholder shared across edges.
This is the case during the offline pretraining of the planner, in which no executor is involved.

\subsection{Backbone with Relative-Pose Attention}\label{app:backbone}

The instruction tokens and the graph tokens are concatenated into one sequence and fused by a Transformer encoder with $L=4$ layers, 12 attention heads and feed-forward size $4d$.
Its self-attention is extended so that the geometric relation between two graph tokens enters the attention computation directly.

\textbf{Poses of tokens.}
Every graph token stands for a place with a pose, attached to a node of the visited tree.
A node token has the pose of its node.
An edge token has the position of its anchor, the approach direction $\hat u_e$ as its heading and the height of its parent, and is attached to the parent node $v_j$.
The STOP token has the pose of the current node.

\textbf{Pairwise descriptor.}
For every ordered pair of graph tokens $(p,q)$ we compute a descriptor $\mathbf r_{pq}\in\mathbb R^9$:
\begin{equation}
\begin{aligned}
\mathbf r_{pq}=\big[&\tfrac{d_{pq}}{d_{\mathrm{norm}}},\ \cos\beta_{pq},\ \sin\beta_{pq},\ \cos\Delta\theta_{pq},\ \sin\Delta\theta_{pq},\\
&\Delta h_{pq},\ \tfrac{\tau_{pq}}{\tau_{\mathrm{norm}}},\ \tfrac{\Delta t_{pq}}{T_{\max}},\ \mathbb 1_{c(p)=c(q)}\big],
\end{aligned}
\label{eq:app-descriptor}
\end{equation}
where $d_{pq}$ and $\beta_{pq}$ are the horizontal distance and the bearing of $q$'s position in $p$'s frame, $\Delta\theta_{pq}$ and $\Delta h_{pq}$ are the differences of their headings and heights, $\tau_{pq}$ is the tree distance between the nodes the two tokens are attached to, that is, the length of the tree path the agent would have to replay to go from one to the other, $\Delta t_{pq}$ is the difference of the macro steps at which the two elements were created, and the last entry indicates whether both tokens are edges that lead to the same frontier.
A two-layer $\mathrm{MLP}$ maps $\mathbf r_{pq}$ to one embedding $\mathbf e_{pq}\in\mathbb R^{d_h}$ per attention head, with $d_h=d/12$.
The embedding is computed once per macro step and shared by all layers.
For pairs in which at least one token is an instruction token, $\mathbf e_{pq}$ is a learned constant.

\textbf{Attention.}
Following~\cite{shaw2018relative}, the embedding enters the keys and the values of every attention head:
\begin{equation}
\begin{aligned}
\alpha_{pq}&=\mathrm{softmax}_q\,\frac{\mathbf q_p^{\top}(\mathbf k_q+W_K\mathbf e_{pq})}{\sqrt{d_h}},\\
\mathbf o_p&=\sum\nolimits_q\alpha_{pq}\,(\mathbf v_q+W_V\mathbf e_{pq}),
\end{aligned}
\label{eq:app-relattn}
\end{equation}
where $\mathbf q_p$, $\mathbf k_q$ and $\mathbf v_q$ are the query, key and value of the tokens and $W_K$, $W_V$ are learned per layer.
The same mechanism covers node--node, node--edge and edge--edge pairs.
It lets the geometric relation interact with what each token is looking for, rather than act as a fixed bias, and it keeps every coordinate out of the tokens, so nothing has to be re-encoded as the agent moves and the graph grows.

\subsection{Heads}\label{app:heads}

Let $\tilde{\mathbf h}_e$, $\tilde{\mathbf h}_{\mathrm{stop}}$ and $\tilde{\mathbf h}_j$ denote the fused tokens output by the last layer.

\textbf{Action head.}
A linear layer scores every fused edge token and the fused STOP token; $\pi_E$ is the softmax over these scores.
Because the edge token contains the value-field encoding $\bar{\mathbf f}_e$, the planner sees the executor's estimates when it chooses the edge.

\textbf{Offset head.}
For the selected edge, an $\mathrm{MLP}$ takes the fused token $\tilde{\mathbf h}_e$ concatenated with the raw field values $[\mathcal F_e(\eta_1),\dots,\mathcal F_e(\eta_{N_F})]$ and outputs the mean $\boldsymbol\mu\in\mathbb R^2$ and the log-variance of a diagonal Gaussian.
A sample $\mathbf z$ of this Gaussian is squashed radially onto the offset disk,
\begin{equation}
\eta=r_{\max}\tanh(\|\mathbf z\|)\,\frac{\mathbf z}{\|\mathbf z\|},
\label{eq:app-squash}
\end{equation}
and the log-density of $\eta$ used by the policy-gradient loss includes the Jacobian of this map.
The offset head is trained only in stage 3, by reinforcement learning; the output layer of $\boldsymbol\mu$ is zero-initialized and the initial log-variance is set small, so that stage 3 starts from a zero offset and widens the distribution only where the reward favors it.
The head sees the raw field values again so that the offset decision can use the shape of the field over the disk, and not only the summary that entered the token.

\textbf{Value head.}
Attention pooling with a learned query over all fused tokens, followed by an $\mathrm{MLP}$, outputs $V_H$.
Because the pooled tokens include the edge tokens, the estimate is aware of the actions currently available, as in the action-aware critic of~\cite{ye2026revisiting}.

\subsection{Executor Network}\label{app:executor}

The executor is a recurrent point-goal policy with the architecture of the DD-PPO PointNav agent~\cite{wijmans2019ddppo}, ResNet-50 followed by a two-layer LSTM with hidden size 512, adapted to the decision space of MATCH.

\textbf{Inputs.}
At decision $k$ the executor receives three inputs.
The forward depth view is encoded by the shared frozen depth encoder into a 512-dimensional feature.
The target is encoded by a linear layer from $[\rho_k/\rho_{\mathrm{norm}},\cos\theta_k,\sin\theta_k]$, where $\rho_k$ and $\theta_k$ are the distance and bearing of the target computed from odometry, into 32 dimensions.
The previous decision is encoded by a learned embedding table with $M+2$ entries, one per heading, one for REACHED and one start symbol used at the first decision, also into 32 dimensions.
The three encodings are concatenated and fed to the LSTM.

\textbf{Outputs.}
Three linear heads on the LSTM state output the REACHED logit, whose sigmoid is $p^R$, the $M$ heading logits, whose softmax is $\pi^{\mathrm{dir}}$, and the value $V_L$.
At test time the executor acts greedily, taking REACHED if $p^R$ exceeds every $(1-p^R)\,\pi^{\mathrm{dir}}(m)$ and the most probable heading otherwise.

\textbf{Initialization.}
The depth encoder with its output projection, the target encoder and the LSTM are initialized from the released DD-PPO checkpoint (ResNet-50 + LSTM-512), whose point-goal input has the same distance-and-bearing form as our target and whose previous-action embedding occupies the same input slot as ours.
The previous-decision embedding table, the two policy heads and the value head are new, because the decision space and the reward of MATCH differ from those of PointNav, and are trained from scratch.
The depth encoder is frozen; the target encoder, the LSTM, the embedding table and the heads are trained in stages 1 and 3.

\textbf{Initial state and value field.}
A sub-task starts with the LSTM state set to zero and the previous decision set to the start symbol.
The first depth view is the view of the start node that faces the anchor, taken from the node's cache, so no additional rendering is needed; from the second decision on, the executor observes the forward view rendered after each move.
The value field of an edge is $V_L$ evaluated in exactly this initial state for the $N_F$ targets $a_e+B_e\eta_n$.
Since the depth feature and the LSTM state are shared by all $N_F$ targets and only the target encoding differs, the field of an edge costs one batched pass of the target encoder, the LSTM and the value head, and the fields of all new edges of a node are computed in a single batch.

\subsection{Modules and Training Status}\label{app:modules}

Table~\ref{tab:app-modules} lists every module, its size and initialization, and whether it is trained in each of the three stages of the main paper.
In total MATCH has 0.5B parameters, most of which are in the frozen encoders and in InfoXLM.

\begin{table}[t]
\caption{Modules of MATCH and their status in the three training stages: $\checkmark$ trained, $\circ$ frozen, -- not used.}
\label{tab:app-modules}
\centering
\footnotesize
\setlength{\tabcolsep}{3pt}
\begin{tabular}{l l l ccc}
\toprule
Module & Size & Init. & S1 & S2 & S3\\
\midrule
\multicolumn{6}{l}{\textit{Shared encoders}}\\
RGB encoder & SigLIP2 ViT-B/16 & pretrained & $\circ$ & $\circ$ & $\circ$\\
Depth encoder & ResNet-50, 512-d & DD-PPO & $\circ$ & $\circ$ & $\circ$\\
\midrule
\multicolumn{6}{l}{\textit{Planner}}\\
Instruction encoder & InfoXLM-base & pretrained & $\checkmark$ & $\checkmark$ & $\checkmark$\\
Panorama encoder & 2 layers, 768 & random & $\checkmark$ & $\checkmark$ & $\checkmark$\\
Token encoders & MLPs, embeddings & random & $\checkmark$ & $\checkmark$ & $\checkmark$\\
Fusion Transformer & 4 layers, 768 & random & $\checkmark$ & $\checkmark$ & $\checkmark$\\
Relative-pose MLP & $9\to64$ per head & random & $\checkmark$ & $\checkmark$ & $\checkmark$\\
Action head & linear & random & $\checkmark$ & $\checkmark$ & $\checkmark$\\
Offset head & MLP & zero & -- & -- & $\checkmark$\\
Value head $V_H$ & pooling + MLP & random & -- & $\checkmark$ & $\checkmark$\\
\midrule
\multicolumn{6}{l}{\textit{Executor}}\\
Target encoder & linear $3\to32$ & DD-PPO & $\checkmark$ & $\circ$ & $\checkmark$\\
Prev.\ decision embedding & $(M{+}2)\times32$ & random & $\checkmark$ & $\circ$ & $\checkmark$\\
LSTM & 2 layers, 512 & DD-PPO & $\checkmark$ & $\circ$ & $\checkmark$\\
Policy heads & linear, $1+M$ & random & $\checkmark$ & $\circ$ & $\checkmark$\\
Value head $V_L$ & linear & random & $\checkmark$ & $\circ$ & $\checkmark$\\
\bottomrule
\end{tabular}
\end{table}

\bibliography{root}
\bibliographystyle{ieeetr}

%% file: root.bbl
\begin{thebibliography}{10}

\bibitem{krantz2020beyond}
J.~Krantz, E.~Wijmans, A.~Majumdar, D.~Batra, and S.~Lee, ``Beyond the nav-graph: Vision-and-language navigation in continuous environments,'' in {\em European Conference on Computer Vision}, pp.~104--120, Springer, 2020.

\bibitem{an2024etpnav}
D.~An, H.~Wang, W.~Wang, Z.~Wang, Y.~Huang, K.~He, and L.~Wang, ``Etpnav: Evolving topological planning for vision-language navigation in continuous environments,'' {\em IEEE Transactions on Pattern Analysis and Machine Intelligence}, 2024.

\bibitem{wei2026dualvln}
M.~Wei, C.~Wan, P.~Peng, X.~Yu, Y.~Yang, D.~Feng, W.~Cai, C.~Zhu, T.~Wang, J.~Pang, {\em et~al.}, ``Ground slow, move fast: A dual-system foundation model for generalizable vision-language navigation,'' in {\em International Conference on Learning Representations}, vol.~2026, pp.~12380--12396, 2026.

\bibitem{ye2026revisiting}
S.~Ye, S.~Mao, Y.~Cui, Y.~Wei, X.~Yu, S.~Zhai, W.~Chen, S.~Zhou, R.~Xiong, and Y.~Wang, ``Revisiting topological graphs for macro action based closed-loop reinforcement learning of vision language navigation in continuous environment,'' {\em arXiv preprint arXiv:2609.03906}, 2026.

\bibitem{zhang2026navfom}
J.~Zhang, A.~Li, Y.~Qi, M.~Li, J.~Liu, S.~Wang, H.~Liu, G.~Zhou, Y.~Wu, X.~Li, {\em et~al.}, ``Embodied navigation foundation model,'' in {\em International Conference on Learning Representations}, vol.~2026, pp.~127293--127322, 2026.

\bibitem{xue2026omninav}
X.~Xue, J.~Hu, M.~Luo, S.~Xie, J.~Chen, Z.~Xie, Q.~Kuichen, W.~Guo, Z.~Chu, M.~Xu, {\em et~al.}, ``Omninav: A unified framework for prospective exploration and visual-language navigation,'' in {\em International Conference on Learning Representations}, vol.~2026, pp.~106216--106231, 2026.

\bibitem{parr1998hmdp}
R.~E. Parr, {\em Hierarchical control and learning for Markov decision processes}.
\newblock University of California, Berkeley, 1998.

\bibitem{hnr}
Z.~Wang, X.~Li, J.~Yang, Y.~Liu, J.~Hu, M.~Jiang, and S.~Jiang, ``Lookahead exploration with neural radiance representation for continuous vision-language navigation,'' in {\em Proceedings of the IEEE/CVF conference on computer vision and pattern recognition}, pp.~13753--13762, 2024.

\bibitem{sutton1999hrl}
R.~S. Sutton, D.~Precup, and S.~Singh, ``Between mdps and semi-mdps: A framework for temporal abstraction in reinforcement learning,'' {\em Artificial intelligence}, vol.~112, no.~1-2, pp.~181--211, 1999.

\bibitem{hong2022bridging}
Y.~Hong, Z.~Wang, Q.~Wu, and S.~Gould, ``Bridging the gap between learning in discrete and continuous environments for vision-and-language navigation,'' in {\em Proceedings of the IEEE/CVF conference on computer vision and pattern recognition}, pp.~15439--15449, 2022.

\bibitem{chen2022duet}
S.~Chen, P.-L. Guhur, M.~Tapaswi, C.~Schmid, and I.~Laptev, ``Think global, act local: Dual-scale graph transformer for vision-and-language navigation,'' in {\em 2022 IEEE/CVF Conference on Computer Vision and Pattern Recognition (CVPR)}, pp.~16516--16526, IEEE, 2022.

\bibitem{an2022bevbert}
D.~An, Y.~Qi, Y.~Li, Y.~Huang, L.~Wang, T.~Tan, and J.~Shao, ``Bevbert: Multimodal map pre-training for language-guided navigation,'' {\em arXiv preprint arXiv:2212.04385}, 2022.

\bibitem{fang2026hsan}
X.~Fang, W.~Fang, and C.~Wang, ``Hierarchical semantic-augmented navigation: Optimal transport and graph-driven reasoning for vision-language navigation,'' {\em Advances in Neural Information Processing Systems}, vol.~38, pp.~61926--61947, 2026.

\bibitem{zhang2020hrac}
T.~Zhang, S.~Guo, T.~Tan, X.~Hu, and F.~Chen, ``Generating adjacency-constrained subgoals in hierarchical reinforcement learning,'' {\em Advances in neural information processing systems}, vol.~33, pp.~21579--21590, 2020.

\bibitem{lee2022dhrl}
S.~Lee, J.~Kim, I.~Jang, and H.~J. Kim, ``Dhrl: A graph-based approach for long-horizon and sparse hierarchical reinforcement learning,'' {\em Advances in Neural Information Processing Systems}, vol.~35, pp.~13668--13678, 2022.

\bibitem{nachum2018hiro}
O.~Nachum, S.~S. Gu, H.~Lee, and S.~Levine, ``Data-efficient hierarchical reinforcement learning,'' {\em Advances in neural information processing systems}, vol.~31, 2018.

\bibitem{levy2017hac}
A.~Levy, G.~Konidaris, R.~Platt, and K.~Saenko, ``Learning multi-level hierarchies with hindsight,'' {\em arXiv preprint arXiv:1712.00948}, 2017.

\bibitem{ahn2022saycan}
M.~Ahn, A.~Brohan, N.~Brown, Y.~Chebotar, O.~Cortes, B.~David, C.~Finn, C.~Fu, K.~Gopalakrishnan, K.~Hausman, {\em et~al.}, ``Do as i can, not as i say: Grounding language in robotic affordances,'' {\em arXiv preprint arXiv:2204.01691}, 2022.

\bibitem{raychaudhuri2021law}
S.~Raychaudhuri, S.~Wani, S.~Patel, U.~Jain, and A.~Chang, ``Language-aligned waypoint (law) supervision for vision-and-language navigation in continuous environments,'' in {\em Proceedings of the 2021 conference on empirical methods in natural language processing}, pp.~4018--4028, 2021.

\bibitem{ilharco2019ndtw}
G.~Ilharco, V.~Jain, A.~Ku, E.~Ie, and J.~Baldridge, ``General evaluation for instruction conditioned navigation using dynamic time warping,'' {\em arXiv preprint arXiv:1907.05446}, 2019.

\bibitem{PPO}
J.~Schulman, F.~Wolski, P.~Dhariwal, A.~Radford, and O.~Klimov, ``Proximal policy optimization algorithms,'' {\em arXiv preprint arXiv:1707.06347}, 2017.

\bibitem{VLN}
P.~Anderson, Q.~Wu, D.~Teney, J.~Bruce, M.~Johnson, N.~S{\"u}nderhauf, I.~Reid, S.~Gould, and A.~Van Den~Hengel, ``Vision-and-language navigation: Interpreting visually-grounded navigation instructions in real environments,'' in {\em Proceedings of the IEEE conference on computer vision and pattern recognition}, pp.~3674--3683, 2018.

\bibitem{RXR}
A.~Ku, P.~Anderson, R.~Patel, E.~Ie, and J.~Baldridge, ``Room-across-room: Multilingual vision-and-language navigation with dense spatiotemporal grounding,'' {\em arXiv preprint arXiv:2010.07954}, 2020.

\bibitem{savva2019habitat}
M.~Savva, A.~Kadian, O.~Maksymets, Y.~Zhao, E.~Wijmans, B.~Jain, J.~Straub, J.~Liu, V.~Koltun, J.~Malik, {\em et~al.}, ``Habitat: A platform for embodied ai research,'' in {\em Proceedings of the IEEE/CVF international conference on computer vision}, pp.~9339--9347, 2019.

\bibitem{qi2025vlnr1}
Z.~Qi, Z.~Zhang, Y.~Yu, J.~Wang, and H.~Zhao, ``Vln-r1: Vision-language navigation via reinforcement fine-tuning,'' {\em arXiv preprint arXiv:2506.17221}, 2025.

\bibitem{wei2025streamvln}
M.~Wei, C.~Wan, X.~Yu, T.~Wang, Y.~Yang, X.~Mao, C.~Zhu, W.~Cai, H.~Wang, Y.~Chen, {\em et~al.}, ``Streamvln: Streaming vision-and-language navigation via slowfast context modeling,'' {\em arXiv preprint arXiv:2507.05240}, 2025.

\bibitem{macenski2022ros}
S.~Macenski, T.~Foote, B.~Gerkey, C.~Lalancette, and W.~Woodall, ``Robot operating system 2: Design, architecture, and uses in the wild,'' {\em Science robotics}, vol.~7, no.~66, p.~eabm6074, 2022.

\bibitem{wen2025foundationstereo}
B.~Wen, M.~Trepte, J.~Aribido, J.~Kautz, O.~Gallo, and S.~Birchfield, ``Foundationstereo: Zero-shot stereo matching,'' in {\em 2025 IEEE/CVF Conference on Computer Vision and Pattern Recognition (CVPR)}, pp.~5249--5260, IEEE, 2025.

\bibitem{macenski2021slam}
S.~Macenski and I.~Jambrecic, ``Slam toolbox: Slam for the dynamic world,'' {\em Journal of Open Source Software}, vol.~6, no.~61, p.~2783, 2021.

\bibitem{fox1999amcl}
D.~Fox, W.~Burgard, F.~Dellaert, and S.~Thrun, ``Monte carlo localization: Efficient position estimation for mobile robots,'' {\em Aaai/iaai}, vol.~1999, no.~343-349, pp.~2--2, 1999.

\end{thebibliography}


\begin{thebibliography}{1}

\bibitem{tschannen2025siglip2}
M.~Tschannen, A.~Gritsenko, X.~Wang, M.~F. Naeem, I.~Alabdulmohsin, N.~Parthasarathy, T.~Evans, L.~Beyer, Y.~Xia, B.~Mustafa, {\em et~al.}, ``Siglip 2: Multilingual vision-language encoders with improved semantic understanding, localization, and dense features,'' {\em arXiv preprint arXiv:2502.14786}, 2025.

\bibitem{wijmans2019ddppo}
E.~Wijmans, A.~Kadian, A.~Morcos, S.~Lee, I.~Essa, D.~Parikh, M.~Savva, and D.~Batra, ``Dd-ppo: Learning near-perfect pointgoal navigators from 2.5 billion frames,'' {\em arXiv preprint arXiv:1911.00357}, 2019.

\bibitem{chi2021infoxlm}
Z.~Chi, L.~Dong, F.~Wei, N.~Yang, S.~Singhal, W.~Wang, X.~Song, X.-L. Mao, H.-Y. Huang, and M.~Zhou, ``Infoxlm: An information-theoretic framework for cross-lingual language model pre-training,'' in {\em Proceedings of the 2021 Conference of the North American Chapter of the Association for Computational Linguistics: Human Language Technologies}, pp.~3576--3588, 2021.

\bibitem{shaw2018relative}
P.~Shaw, J.~Uszkoreit, and A.~Vaswani, ``Self-attention with relative position representations,'' in {\em Proceedings of the 2018 Conference of the North American Chapter of the Association for Computational Linguistics: Human Language Technologies, Volume 2 (Short Papers)}, pp.~464--468, 2018.

\bibitem{ye2026revisiting}
S.~Ye, S.~Mao, Y.~Cui, Y.~Wei, X.~Yu, S.~Zhai, W.~Chen, S.~Zhou, R.~Xiong, and Y.~Wang, ``Revisiting topological graphs for macro action based closed-loop reinforcement learning of vision language navigation in continuous environment,'' {\em arXiv preprint arXiv:2609.03906}, 2026.

\end{thebibliography}
